\def\year{2021}\relax
\documentclass[letterpaper]{article} % DO NOT CHANGE THIS
\usepackage{aaai21}  % DO NOT CHANGE THIS
\usepackage{times}  % DO NOT CHANGE THIS
\usepackage{helvet} % DO NOT CHANGE THIS
\usepackage{courier}  % DO NOT CHANGE THIS
\usepackage[hyphens]{url}  % DO NOT CHANGE THIS
\usepackage{graphicx} % DO NOT CHANGE THIS
\usepackage{natbib}  % DO NOT CHANGE THIS AND DO NOT ADD ANY OPTIONS TO IT
\usepackage{caption} % DO NOT CHANGE THIS AND DO NOT ADD ANY OPTIONS TO IT
\usepackage{graphicx}
\usepackage{comment}
\usepackage{amsmath,amssymb}
\usepackage{color}
\usepackage{hyperref}
\usepackage{bm}
\usepackage{amssymb}
\usepackage{amsfonts,amssymb}
\usepackage{gensymb}
\usepackage{float}
\usepackage{algorithm}
\usepackage{algorithmic}
\usepackage{verbatim}
\usepackage[switch]{lineno}

\DeclareMathOperator*{\argmax}{arg\,max}
\newcommand{\tabincell}[2]{\begin{tabular}{@{}#1@{}}#2\end{tabular}}

\title{VMLoc: Importance Variational Multimodal Fusion \\ for Localization}

\author{Kaichen Zhou, Changhao Chen*, Bing Wang, \\ Muhamad Risqi U. Saputra, Niki Trigoni, Andrew Markham}

\affiliations{ \\
Department of Computer Science, University of Oxford\\
$\{$rui.zhou, changhao.chen, bing.wang, muhamad.saputra, niki.trigoni, andrew.markham$\}$@cs.ox.ac.uk\\}

\begin{document}
%\linenumbers
\newcommand{\cch}[1]{{\color{magenta}{Changhao: #1}}}
\newcommand{\acm}[1]{{\color{red}{Andrew: #1}}}
\newcommand{\bw}[1]{{\color{green}{Bing WANG: #1}}}
\newcommand{\al}[1]{{\color{blue}{Alex97: #1}}}
\newcommand{\ri}[1]{{\color{cyan}{Risqi: #1}}}
\maketitle

\begin{abstract}
Recent learning-based approaches have achieved impressive results in the field of single-shot camera localization. However, how best to fuse multiple modalities (e.g., image and depth) and to deal with degraded or missing input are less well studied. In particular, we note that previous approaches towards deep fusion do not perform significantly better than models employing a single modality. We conjecture that this is because of the naive approaches to feature space fusion through summation or concatenation which do not take into account the different strengths of each modality. To address this, we propose an end-to-end framework, termed VMLoc, to fuse different sensor inputs into a common latent space through a variational Product-of-Experts (PoE) followed by attention-based fusion. Unlike previous multimodal variational works directly adapting the objective function of vanilla variational auto-encoder, we show how camera localization can be accurately estimated through an unbiased objective function based on importance weighting. Our model is extensively evaluated on RGB-D datasets and the results prove the efficacy of our model. The source code is available at \href{https://github.com/kaichen-z/VMLoc}{https://github.com/kaichen-z/VMLoc}.
\end{abstract}

\section{Introduction}
Visual localization is of great importance to many intelligent systems, e.g. autonomous vehicles, delivery drones, and virtual reality (VR) devices. Recently, deep learning has shown its strengths in learning camera pose regression from raw images in an end-to-end manner. However, a single modality solution is normally confronted with issues such as environmental dynamics, changes in lighting conditions, and extreme weather, when it is deployed in complex and ever-changing real-world environments. Meanwhile, those intelligent systems are generally equipped with a combination of sensors (e.g. RGB cameras, depth sensors, and LIDAR) which can be exploited to improve the robustness of the systems. Nevertheless, most studies concentrating on sensor fusion \cite{liang2019multi,bijelic2019seeing} directly concatenates different feature vectors together without studying the information contained in different feature vectors. Effectively exploiting different sensor modalities and studying their complementary features will contribute to a more accurate and robust localization system. 

\begin{figure}
\setlength{\abovecaptionskip}{0.cm}
\setlength{\belowcaptionskip}{-0.5cm}
    \centering
    \includegraphics[width=3.3in]{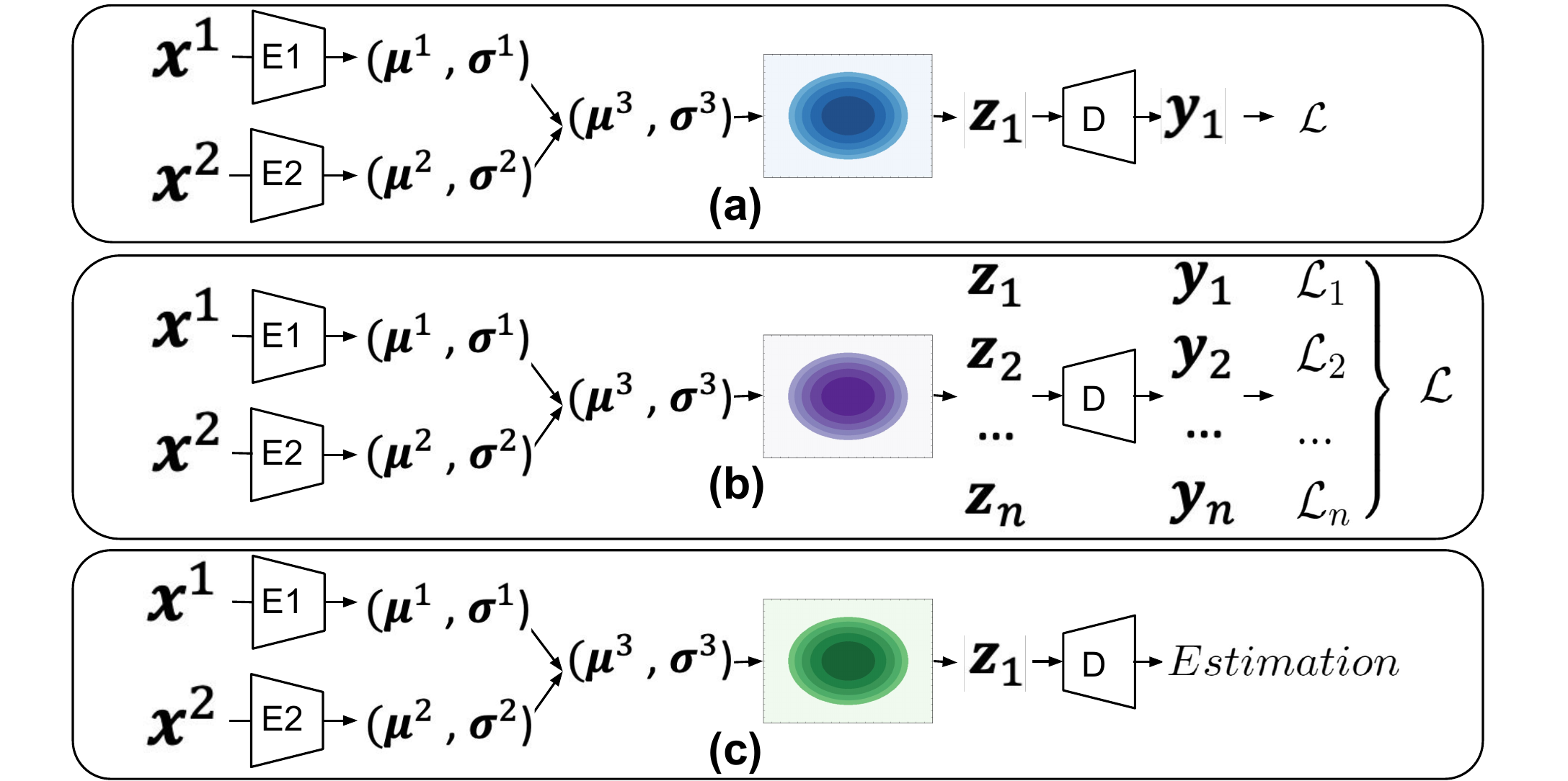}
    \caption{(a) Training process of MVAE. (b) Training process of VMLoc. (c) Inference process of VMLoc.}
    \label{fig:multimodal_fusion}
\end{figure}

Compared with learning-based localization algorithms using only RGB images, e.g. PoseNet \cite{kendall2015posenet}, Bayesian PoseNet \cite{kendall2016modelling}, Hourglass Network \cite{melekhov2017image}, CNN+LSTM localization neural network \cite{walch2017image}, PoseNet17 \cite{kendall2017geometric}, etc., multimodal localization is far less to be explored. VidLoc \cite{clark2017vidloc} simply concatenates RGB image and depth map, and processes them together with Convolutional Neural Networks (CNN) for camera localization. However, this direct concatenation on the raw data level won't take advantage of the complementary properties of different modalities. MVAE \cite{wu2018multimodal} proposes to use variational inference to learn an invariant space from multimodal fusion which achieved impressive results in several tasks. This is done by directly optimizing the evidence lower bound (ELBO) whose representation is simplified \cite{burda2015importance}. However, MVAE is hard to scale to real-world environments with high-dimensional raw images and rich scene information (e.g. for visual localization problem) as the objective function of MVAE cannot give a tighter estimation of ELBO \cite{burda2015importance}. 

In order to address the aforementioned problems, we propose VMLoc (\textbf{V}ariational Fusion For \textbf{M}ultimodal Camera \textbf{Loc}alization), a novel framework to learn multimodal 6-DoF camera localization, which learns a joint latent representation from a pair of sensor modalities (as depicted in Figure \ref{fig:multimodal_fusion} (b) and (c)). Our intuition is that there should exist a common space from multiple modalities that is useful and suitable for solving the task at hand. By using Product-of-Experts (PoE), we combine the individual latent spaces of each modality while at the same time enforcing each modality to concentrate on the specific property that is more useful for the task. We then propose unbiased objective function based on importance weighting to provide the framework with a tighter estimation of ELBO. The main contributions of this work can be summarized as follow:
\begin{itemize}
\item We introduce VMLoc, a novel deep neural network framework to combine a pair of sensor modalities, e.g. vision and depth/lidar, for camera localization problem.
\item We propose a PoE fusion module to learn the common latent space of different modalities by using an unbiased objective function based on importance weighting.
\item Extensive experiments on indoor and outdoor scenarios and systematic research into the robustness and ablation demonstrate the effectiveness of our proposed framework. 
\end{itemize}

\section{Related Work}
\subsection{Sensor Fusion and Multimodal learning}
Because of the complementary properties of different sensors, an effective and suitable fusion strategy plays a vital role in learning from multiple sensor modalities in diverse fields, e.g., \cite{misra2016cross}, \cite{valada2019self}, and \cite{mees2016choosing}.

However, to date, limited research has considered sensor fusion in the context of visual localization, e.g., \cite{biswas2013multi} and \cite{chen2019selective}. Nevertheless, existing frameworks have very limited performances and they could not provide a better result than state-of-the-art approaches based on monomodal.
%\cite{biswas2013multi} analyzes localization performance of the laser rangefinder, the depth camera, and the WiFi. They identify the sensor which can provide the most accurate location estimation in different situations. \cite{chen2019selective} proposes an end-to-end selective fusion framework for monocular VIO which fuses monocular images and inertial measurements to estimate the trajectory. Nevertheless, their performances are still very limited and they could not provide better result than state-of-the-art approaches based on monomodality. 
Learning a joint representation and studying the respective contribution from different modalities are also the focus of multi-modal learning. Among them, some works propose to learn an explicit joint distribution of all modalities, e.g., Joint Multi-modal Variational Auto-encoder (JMVAE) \cite{suzuki2016joint} and PoE \cite{wu2018multimodal}. However, both JMVAE and PoE are only tested in simple, simulated datasets, such as MNIST \cite{lecun1998gradient} or Fashion MNIST \cite{xiao2017fashion}.
%Their performances are doubtful when dealing with more complicated real-life tasks. \cite{suzuki2016joint} proposes a joint multi-modal variational auto-encoder (JMVAE) to extract a joint representation among all models, in which all modalities are independently conditioned on the joint representation. \cite{wu2018multimodal} uses a PoE inference network and a sub-sampled training paradigm to learn a joint representation.
Other works propose to learn the individual subspace and to achieve cross inference among modalities, e.g., mixture-of-experts (MoE) multimodal variational autoencoder (MMVAE) \cite{shi2019variational} and symbol-concept association network (SCAN) \cite{higgins2017scan}. Compared with them, our work introduces importance weighting strategy into the multimodal variational model to improve its modelling capacity and to reduce the training variance. Meanwhile, we incorporate the geometric loss into the inference process to encourage useful features for pose estimation.

%For example, \cite{shi2019variational} proposes a mixture-of-experts (MoE) multimodal variational autoencoder (MMVAE) to learn the individual latent space of each modality and realizes the cross-modal generation. \cite{higgins2017scan} introduces the symbol-concept association network (SCAN) which minimizes the KL divergence between different latent spaces and deals with the multimodal bi-directional inference task. Compared with them, our work introduces importance weighting strategy into the multimodal variational model to improve its modelling capacity and to reduce the training variance. Meanwhile, we incorporate the geometric loss into the inference process to encourage useful features for pose estimation.

\subsection{Camera Localization}
Camera localization methods can be categorized into the conventional, structure-based models \cite{sattler2015hyperpoints,sattler2016efficient,cavallari2019real,brachmann2017dsac,brachmann2018learning} and deep learning-based models \cite{kendall2015convolutional,walch2017image, ding2019camnet,zhou2023manydepth2,zhou2022devnet}. 

\subsubsection{Conventional Approaches}
Conventional, structure-based camera localization typically employs Perspective-n-Point (PnP) algorithm \cite{hartley2003multiple} applied on 2D-to-3D correspondences. The algorithm consists of finding the correspondences between image features (2D) and world space points (3D) and generating multiple camera pose hypotheses based on the correspondences. The outliers rejection method is then used to cull the multiple hypotheses into a single, robust estimation. \cite{williams2011automatic} employed randomized lists classifier to recognize correspondences and use them along with RANSAC to determine the camera pose. \cite{sattler2015hyperpoints} finds locally unique 2D-3D matches and employed them to generate reliable localization based on image retrieval techniques.  \cite{sattler2016efficient} emphasized that it is important to fuse 2D-to-3D and 3D-to-2D search to obtain accurate localization. However, structure-based models usually require large memory and space to save the 3D model and the descriptors \cite{sattler2015hyperpoints}. This leads to the development of efficient "straight-to-pose" methods based on a machine learning algorithm.

\subsubsection{Deep Learning-based Approaches}
Instead of basing on 3D-geometry theory to tackle the problem \cite{chen2011city,guzman2014multi,zeisl2015camera}, deep learning models directly learn useful features from raw data \cite{radwan2018vlocnet++,walch2017image} to regress 6-DoF pose, e.g., PoseNet \cite{kendall2015posenet}, PoseNet++ \cite{melekhov2017image}. Further works propose to implement new constraint or new neural network structure to improve the performance, e.g., \cite{brachman-schmolze:kl-one}, \cite{kendall2017geometric}, \cite{clark2017vinet} and \cite{walch2017image}. To date, deep localization has largely considered a single modality as input. However, our work proposes a fusion framework that can effectively use the data from several different sensors to solve the camera localization problem.

%PoseNet \cite{kendall2015posenet} first leveraged CNN to learn and predict camera poses from single images. Subsequently, the PoseNet structure was further developed to enhance the performance, e.g. by using the skip connection introduced in \cite{melekhov2017image}. Besides architectural refinements, the loss function used in the optimization can also be enhanced. 
%Instead of using the combination of position and rotation errors in the training process \cite{kendall2015convolutional,walch2017image}, visual odometry constraints \cite{brachman-schmolze:kl-one} are incorporated to improve the model performance; \cite{kendall2017geometric} proposed using a learning weighted loss and a geometric re-projection loss to generate the more accurate result. Furthermore, the use of recurrent neural network has been introduced to improve the localization accuracy from both the temporal aspect \cite{clark2017vinet} and the spatial aspect \cite{walch2017image}. 

\section{Methodology}
\subsection{Problem Formulation}
This work is aimed at exploiting multimodal data to achieve more robust and accurate pose estimation $\mathbf{y}=(\mathbf{p},\mathbf{q})$, which consists of a location vector $\mathbf{p} \in \mathbb{R}^3$ and a quaternion based orientation vector $\mathbf{q} \in \mathbb{R}^3$. The multimodal data are the different observations of an identical scene, but complementary to each other. For example, RGB images contain the appearance and the semantic information of the scene, while depth maps or point cloud data capture the scene structure. Intuitively, these sensor modalities reflect a spatial sense of the scene, and hence a common feature space that is useful for solving the task at hand should exist.

Given two sensor modalities $\mathbf{x^1}$ and $\mathbf{x^2}$, we aim to learn their joint latent representation $\mathbf{z}$. As shown in Figure \ref{fig:multimodal_fusion}, this problem is formulated as a Bayesian inference model, which is to maximize the posterior probability conditioned on input data:
\begin{equation}
    \mathbf{z} = \argmax_{\mathbf{z}} [p(\mathbf{z}|\mathbf{x^1}, \mathbf{x^2})].
\end{equation}
Based on this intermediate representation $\mathbf{z}$, the target value $\mathbf{y}$ is obtained by     
\begin{equation}
    \mathbf{y} = \argmax_\mathbf{y} [p(\mathbf{y}|\mathbf{z})].
\end{equation}
The problem becomes how to recover the joint distribution of two modalities. Our work mainly considers a pair of modalities, although the proposed method can be naturally extended to the usage of three or more modalities.   

\subsection{VMLoc}
We introduce VMLoc framework to tackle this multimodal learning problem. Our proposed method leverages the variational inference models \cite{kingma2013auto} to find a distribution $q_{\phi}(\mathbf{z}|\mathbf{x^1}, \mathbf{x^2})$, approximating the true posterior distribution $p(\mathbf{z}|\mathbf{x^1}, \mathbf{x^2})$ with the aid of the corresponding geometric information. Algorithm \ref{al:algo} demonstrates the detailed algorithmic description of our proposed VMLoc.

\begin{figure*}
\setlength{\abovecaptionskip}{0.cm}
\setlength{\belowcaptionskip}{-0.5cm}
    \centering
    \includegraphics[width=5.8in]{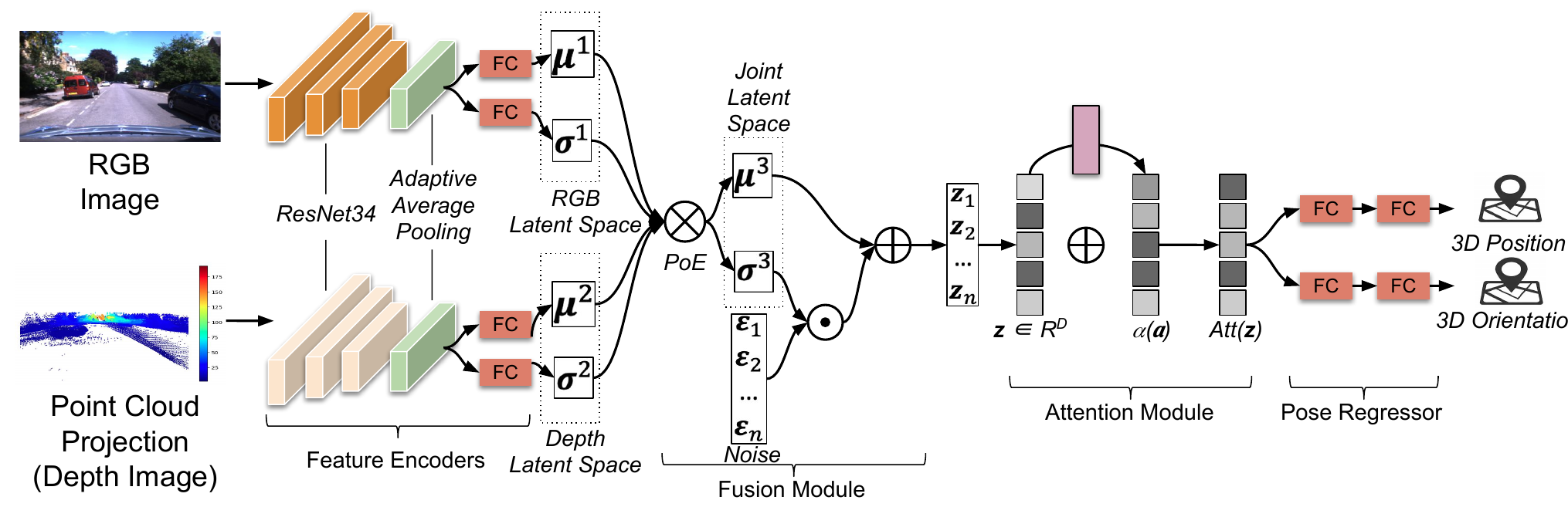}
    \caption{Our framework consists of feature encoders, a fusion module, an attention mechanism module and a pose regressor.}
    \label{fig:fig1}
\end{figure*}

\subsubsection{Multimodal Variational Inference}
Unlike conventional Variantional Auto-Encoders (VAE) \cite{kingma2013auto}, VMLoc reconstructs a target value $\mathbf{y}$ in a different domain, i.e. 6-DoF pose in our case, instead of the original domain data.
Similar to the cross-modality VAE \cite{yang2019aligning}, VMLoc produces the target value $\mathbf{y}$ via a joint latent variable $\mathbf{z}$ of two modalities $\mathbf{x^1}$ and $\mathbf{x^2}$, by maximizing the ELBO as follow: % \cch{Are you sure about "maximize ELBO"? Solved}:
\begin{equation}
\begin{split}
    \log p(\mathbf{y}) & \geq \mathbb{E}_{z \sim q_{\phi}(\mathbf{z}|\mathbf{x^1},\mathbf{x^2})}[\log \frac{p_{\theta}(\mathbf{y}|\mathbf{z})p(\mathbf{z})}{q_{\phi}(\mathbf{z}|\mathbf{x^1},\mathbf{x^2})}] \\
    & = \mathbb{E}_{z \sim q_{\phi}(\mathbf{z}|\mathbf{x^1},\mathbf{x^2})}[p_{\theta}(\mathbf{y}|\mathbf{z})] - \text{KL}(q_{\phi}(\mathbf{z}|\mathbf{x^1},\mathbf{x^2})|p(\mathbf{z}))\\ 
    & = \text{ELBO}(\mathbf{y};\mathbf{x^1},\mathbf{x^2}) \\
\end{split}
\end{equation}
where $p(\mathbf{y})$ is the distribution of target value, $p(\mathbf{z})$ is the prior distribution of latent space, $q_{\phi}(\mathbf{z}|\mathbf{x^1},\mathbf{x^2})$ is the inference model to approximate the posterior distribution $p(\mathbf{z}|\mathbf{x^1},\mathbf{x^2})$, and $p_{\theta}(\mathbf{y}|\mathbf{z})$ is the decoder network.

Inspired by MVAE \cite{wu2018multimodal}, the inference problem of $p(\mathbf{z}|\mathbf{x^1},\mathbf{x^2})$ can be simplified as learning two conditional distributions $p(\mathbf{z}|\mathbf{x^1})$ and $p(\mathbf{z}|\mathbf{x^2})$ separately. This is under the assumption that two modalities are conditionally independent given the latent representation. 
Based on that, we can apply the Product-of-Experts (PoE) technique to estimate the joint latent distribution $p(\mathbf{z}|\mathbf{x^1},\mathbf{x^2})$ via: 
\begin{equation}
    p(\mathbf{z}|\mathbf{x^1},\mathbf{x^2}) = \frac{p(\mathbf{z}|\mathbf{x^1})p(\mathbf{z}|\mathbf{x^2})}{p(\mathbf{z})}.
    \label{equ: poe}
\end{equation}
Here, PoE works to combine several simple distributions by producing their density functions \cite{hinton1999products}. In our case, it allows each modality to specialize in their specific property to contribute to the final pose estimation, rather than forcing each modality to recover the full-dimensional information to solve the problem.

As the distribution $p(\mathbf{z}|\mathbf{x^1})$ can be approximated with an inference network $q(\mathbf{z}|\mathbf{x^1}) \equiv \tilde{q}(\mathbf{z}|\mathbf{x^1}) q(\mathbf{z})$, Equation \ref{equ: poe} is further developed as 
\begin{equation}
p(\mathbf{z}|\mathbf{x^1},\mathbf{x^2}) = \tilde{q}(\mathbf{z}|\mathbf{x^1}) \tilde{q}(\mathbf{z}|\mathbf{x^2}) q(\mathbf{z}).
\label{equ: poe2}
\end{equation}
Thus the learning process can be summarized as: it first learns two individual latent distributions $p(\mathbf{z}|\mathbf{x^1})$ and $p(\mathbf{z}|\mathbf{x^2})$; then combines two distributions to obtain the final joint distribution. In practice, we assume that both the prior distribution and the posterior distribution are Gaussian distributions. With the mean vector and the variance matrix, we can sample from the learned distribution to obtain the joint latent representation of the input modalities.

\subsubsection{Importance Weighting Strategy}
However, directly optimizing $\text{ELBO}$ as the objective function can only provide a simplified representation. %\cch{Why it leads to simplified representation? use the figure to describe the limitation of single resampling vs our approach multiple resampling? This part is your contribution and important.}\cch{Briefly explain. That is the reason why you improve mvae.} \al{I will add a figure.} 
As shown in Figure \ref{fig:multimodal_fusion} (a), in the normal variational model, e.g. MVAE, the representation is only re-sampled once from the latent distribution. In this way, the capability of variational inference has not been fully exploited and our latter experiments in Table \ref{table:Ablation} support this argument. Therefore, we introduce the importance weighting strategy into the framework. As Figure \ref{fig:multimodal_fusion} (b) illustrates, instead of only depending on one sample, this strategy resamples from learned distribution for multiple times to approximate the posterior distribution, which allows the network to model more complicated posterior distribution. In doing so, it provides a strictly tighter log-likelihood lower bound \cite{burda2015importance} through:
\begin{equation}
 \mathbb{E} [\log \frac{1}{k} \sum_{i=1}^k w_i] \geq \mathbb{E} [\log \frac{1}{k-1} \sum_{i=1}^{k-1} w_i],
\end{equation}
where $w_i = p(\mathbf{y}, \mathbf{z_i})/q(\mathbf{z_i}|\mathbf{x^1},\mathbf{x^2})$; $k$ is the number of samples; $\mathbf{z_i}$ is the $i$-th point sampled independently from the latent distribution. Thus, the objective function is rewritten as:
\begin{equation}
\begin{split}
   \log p(\mathbf{y})& = \log \mathbb{E} [w] \\
   & \geq \mathbb{E} [\log \frac{1}{k} \sum_{i=1}^k w_i] \\
   & = \text{ELBO}(\mathbf{y};\mathbf{x^1},\mathbf{x^2}).
\label{equ:imae}
\end{split}
\end{equation}
It can be noticed from the above analysis that the larger number of samples leads to a tighter log-likelihood lower bound. With this new objective function, our model uses multiple samples instead of one to approximate the posterior, which then improves its capability to learn more complex posterior.
Although by increasing the number of samples we can find a tighter bound, it counter-intuitively leads to a higher variance for gradient estimation in the training process \cite{rainforth2018tighter}. The gradient for $\bm{\phi}$ in Equation \ref{equ:imae} are calculated via:
\begin{equation}
\begin{split}
       \bigtriangledown_{\bm{\phi}} 	L_k & = \mathbb{E}_{\varepsilon_{1:k}} [\sum_{i=1}^k \frac{w_i}{\sum_j w_j} (\frac{\partial \log w_i}{\partial \mathbf{z_i}}\frac{\partial \mathbf{z_i}}{\partial \bm{\phi}} \\
       & - \frac{\partial}{\partial \bm{\phi}} \log q_{\bm{\phi}}(\mathbf{z_i}|\mathbf{x^1},\mathbf{x^2}))].
\label{equ:gradientinf2}
\end{split}
\end{equation}

According to \cite{rainforth2018tighter}, the second item (score function) in Equation \ref{equ:gradientinf2} can be eliminated, so as to reduce the variance of the gradients. However, they fail to show that eliminating this item is unbiased. To address this concern, \cite{tucker2018doubly} eliminates the score function by rewriting Equation \ref{equ:gradientinf2} as:
\begin{equation}
       \bigtriangledown_{\phi} 	L_k = \mathbb{E}_{\varepsilon_{1:k}} [\sum_{i=1}^k (\frac{w_i}{\sum_j w_j})^2 \frac{\partial\log w_i}{\partial \mathbf{z_i}} \frac{\partial\mathbf{z_i}}{\partial \bm{\phi}} ].
\label{equ:gradientinf3}
\end{equation}
Through Equation \ref{equ:gradientinf3}, we mitigate the side effects of increasing $k$ and ensure the gradients of the inference network are unbiased. Instead of directly computing $w_i = p(\mathbf{y}, \mathbf{z_i})/q(\mathbf{z_i}|\mathbf{x^1},\mathbf{x^2})$ as in Equation \ref{equ:gradientinf3}, in practice, we rewrite it as follow:
\begin{equation}
\begin{split}
w_i & = e^{\log w_i} \\ 
& = e^{\log p(\mathbf{y}|\mathbf{z_i}) + \log p(\mathbf{z_i}) - \log q(\mathbf{z_i}|\mathbf{x^1},\mathbf{x^2})}
\label{equ:w_i}
\end{split}
\end{equation}
We further replace the first logarithm likelihood item in Equation \ref{equ:w_i} with the negative geometric loss which calculates the distance between the predicted pose and the target pose in the pose estimation task. 

%%%%%%%%%%%%%%%%%%%%%%%%%%%%%%%%%%%%%%%%%%%%%%END%%%%%%%%%%%%%%%%%%%%%%%%%%%%%%%%%%%%%%%%%%%%%%%%%%%%%
\subsubsection{Geometric Learning}
To encourage VMLoc to extract useful features for pose estimation, we propose to incorporate geometric information into the optimization objective. We incorporate a learnable geometric loss \cite{clark2017vinet,brahmbhatt2018geometry} into the loss function:
\begin{equation}
    L_p(\mathbf{y},\mathbf{y}^*) = ||\mathbf{p}-\mathbf{p}^*||e^{-\beta} + \beta + || \log \mathbf{q} - \log \mathbf{q}^*||e^{-\gamma} + \gamma 
\label{equ:learnableloss}
\end{equation}
where $\mathbf{y}^*=(\mathbf{p}^*, \mathbf{q}^*)$ is the ground truth pose, while $\beta$ and $\gamma$ are weights that balance the position and the rotation loss. The $\beta$ and the $\gamma$ are optimized during the training process with the initial value $\beta_0$ and $\gamma_0$ as in \cite{zhou2021dha}. The $\log \bm{q}$ is the logarithmic form of a unit quaternion $\bm{q}$, which is defined as:
\begin{equation}
    \log \bm{q} = \left\{
                    \begin{aligned}
                    \frac{\bm{v}}{||\bm{v}||}\cos^{-1} u, & \quad if ||\bm{v}|| \neq 0 \\
                    \bm{0}, & \quad \text{otherwise}
                    \end{aligned}
                    \right.
\end{equation}
where $u$ denotes the real part of a unit quaternion and $\bm{v}$ is the imaginary part \cite{wang2019atloc}. In general, the gradient for the inference network $\bm{\phi}$ and the decoder network $\bm{\theta}$ (also known as the pose regression network) can be expressed as follow:

\begin{equation}
 \bigtriangledown_{\bm{\phi}} L_k = \mathbb{E}_{\varepsilon_{1:k}} [\sum_{i=1}^k (\frac{w_i}{\sum_j w_j})^2 \frac{\partial\log w_i}{\partial \mathbf{z_i}} \frac{\partial \mathbf{z_i}}{\partial \bm{\phi}} ]; 
 \label{equ:gradient_phi}
\end{equation}

\begin{equation}
  \bigtriangledown_{\bm{\theta}} L_k  = \mathbb{E}_{\varepsilon_{1:k}} [\sum_{i=1}^k \frac{w_i}{\sum_j w_j} \bigtriangledown_{\bm{\theta}}\log w_i];  
 \label{equ:gradient_theta}
\end{equation}

\begin{equation}
 w_i  = e^{ - L_p(\mathbf{y},\mathbf{y}^*) + \lambda( \log p(\mathbf{z_i}) - \log q(\mathbf{z_i}|\mathbf{x^1},\mathbf{x^2}))};
 \label{equ:gradient_explain}
\end{equation}

where $\lambda$ is the hyperparameter introduced by \cite{higgins2017beta} to balance the prediction accuracy and the latent space capability. The complete learning algorithm can be found in Algorithm \ref{al:algo}.

\begin{algorithm}
	\caption{\quad VMLoc algorithm}
	\par Require: $\mathbf{x^1}$, $\mathbf{x^2}$ and $\mathbf{y}$
	%\par Ensure: $\bm{\phi}_{\mathbf{x^1}}$, $\bm{\phi}_{\mathbf{x^2}}$, $\bm{\theta}$, $\beta$ and $\gamma$ 
	\begin{algorithmic}
	\STATE  Initialize parameters $\bm{\phi_{\mathbf{x^1}}}$, $\bm{\phi_{\mathbf{x^2}}}$, $\bm{\theta}$, $\beta$ and $\gamma$
	\FOR{episode=\(1, N\)} 
    \STATE  Encode $\mathbf{x^1}$ and $\mathbf{x^2}$ with $q_{\bm{\phi}_{\mathbf{x^1}}}(\mathbf{z}|\mathbf{x^1})$ and $q_{\bm{\phi}_{\mathbf{x^2}}}(\mathbf{z}|\mathbf{x^2})$
	\STATE Compute the joint distribution via Equation \ref{equ: poe2}
	\STATE Sample $k$ points $\mathbf{z_i}$ from joint distribution
	\STATE Decode $\mathbf{z_i}$  with $p_{\bm{\theta}}(\mathbf{y}|\mathbf{z})$
	\STATE Update the parameters $\beta$ and $\gamma$ with loss function $L_p(\mathbf{y},\mathbf{y}^*)$ in Equation \ref{equ:learnableloss}
	
	\STATE Update the parameters $\bm{\phi_{\mathbf{x^1}}}$, $\bm{\phi_{\mathbf{x^2}}}$ with gradient $\bigtriangledown_{\bm{\phi}} L_k$ in Equation \ref{equ:gradient_phi}
	
	\STATE Update the parameters  $\bm{\theta}$ with gradient $\bigtriangledown_{\bm{\theta}} L_k$ in Equation \ref{equ:gradient_theta} 
	
    \ENDFOR
    \end{algorithmic}
    \label{al:algo}
\end{algorithm}

\subsection{Framework}
Now we come to introduce the detailed framework and the training strategy of VMLoc. Figure \ref{fig:fig1} illustrates the structure of proposed VMLoc, including an RGB image encoder, a depth map encoder, a fusion module, an attention module, and a pose regressor. These two encoders separately encode the RGB images and the depth maps into their own latent space, followed by fusing the multiple individual latent spaces into one joint latent space through PoE. Then, the latent representation sampled from the joint latent space is re-weighted by a self-attention module. Finally, the re-weighted latent features are taken as the input for the successive pose regressor to predict the 6-DoF camera pose. 

\subsubsection{Feature Encoders}
The feature encoders in our framework include an image encoder $q_{\phi_{img}}(\mathbf{z}|\mathbf{x_{img}})$ and a depth encoder $q_{\phi_{dep}}(\mathbf{z}|\mathbf{x_{dep}})$, which separately learn the latent distribution of RGB image and that of depth map. In learning point cloud feature, we transform the lidar data into a image using the cylindrical projection \cite{chen2017multi}. CNN has already shown its strengths in the task of visual localization \cite{brahmbhatt2018geometry}. Among them, the ResNet model has been widely applied in different tasks, e.g., \cite{brahmbhatt2018geometry,wang2019atloc}. Based on these concerns, in our model, we also adopt the ResNet34 structure to construct our RGB image encoder and depth map encoder. To accelerate the convergence speed, the ResNet34 in our model was initialized by the weights of model trained on the Image-Net \cite{he2016deep}. 
%For the depth map encoder, the first convolution layer with three channels can be replaced by the convolution layer with one channel according to the dimension of depth map. 
For both RGB image encoder and depth map encoder, the second last average pooling layer is replaced by the adaptive average pooling layer and is followed by two parallel fully connected layers with the same dimension $D=1024$, which separately output the mean vector $\bm{\mu}\in\mathbb{R}^D$ and the diagonal vector of the variance matrix $\bm{\sigma}\in\mathbb{R}^D$ of the learned latent distribution.

\subsubsection{Fusion Module} 
After we learned the mean vector and the variance matrix for the latent distribution of the RGB image and the depth map, we compute corresponding parameters for the joint distribution via Equation \ref{equ: poe2}. Rather than directly sampling from the joint distribution which isn't differentiable, the reparameterization trick \cite{kingma2013auto} is applied. Given the mean vector $\bm{\mu}$ and the variance matrix $\bm{\sigma}$, we first sample the noise $\bm{\epsilon} \sim \mathcal{N} (\mathbf{0}, \mathbf{I})$. Then the point of joint latent distribution can be computed as $\mathbf{z}=\bm{\epsilon}\bm{\mu} + \bm{\sigma}$. 

%\cch{This part is not your contribution. Try to reduce it.}

\subsubsection{Attention Module}
Considering that certain parts of the features extracted by the model may be useless to the pose regression, we would like to enable our framework to focus on certain representations that are useful to the task. % and hence improve the generalization capacity of the model. %, which can even cause huge difference for the performance of the model over the training set and the test set. 
We implement the non-local style self-attention \cite{wang2018non} in our attention module, which can capture the long-range dependencies and global correlation of the image features \cite{wang2019atloc}. The computation process of attention module can be summarized as the following two steps. Given a feature vector $\bm{z}\in \mathbb{R}^D$, we firstly calculate its self-attention as follow: 
\begin{equation}
    \bm{a} = \text{Softmax}(\bm{z}^T \bm{W}_{\theta}^T \bm{W}_{\phi} \bm{z}) \bm{W}_{g} \bm{z},
\end{equation}
where $\bm{W}_{\theta}$, $\bm{W}_{\phi}$ and $\bm{W}_g$ are the learn-able weights. Then, the residual connection will be added to the linear embedding of the self-attention vectors:
\begin{equation}
    \text{Att}(\bm{z}) = \alpha(\bm{a}) + \bm{z},  
\end{equation}
where the $\alpha(\bm{a})=\bm{W}_{\alpha}\bm{a}$ and the $\bm{W}_{\alpha}$ is a learnable weight which will be optimized during the training process.

\subsubsection{Pose Regressor}
Finally, the re-weighted latent vector is taken as input into the pose regressor to estimate 6-DoF pose. The pose regressor consists of two parallel networks sharing the same structure. Each network contains two successive fully connected layers connected by a ReLU activation function. Among them, one network predicts the position vector $\mathbf{p}$, while another network predicts the quaternion based rotation vector $\mathbf{q}$.

\subsubsection{Training Strategies}
\begin{comment}
For stability, the training process is divided into two steps. Firstly, we optimize the model via the loss function without the KL divergence part. Correspondingly, in this step, there is no sampling process for the latent distribution and the mean of the distribution is directly used in the following position regression. In the second step, we train the model with the complete loss function. Beside the training strategy,
\end{comment}
In order to force our model to learn from all input modalities and to improve the robustness during the corrupted input conditions, a data augmentation method for multimodal learning is introduced. This data augmentation method can be formulated as follow:
\begin{equation}
    (\mathbf{x^1}, \mathbf{x^2}) = \left\{
                    \begin{aligned}
                    (\mathbf{x^1}, \mathbf{x^2}), & \quad with\quad p_1 \\
                    (\mathbf{x^1}, None), & \quad with\quad p_2 \\
                    (None, \mathbf{x^2}), & \quad with\quad p_3,
                    \end{aligned}
                    \right.
\end{equation}
where the sum of $p_1$, $p_2$ and $p_3$ is $1$. In our experiments, $p_1=\frac{3}{5}$, $p_2=\frac{1}{5}$ and $p_3=\frac{1}{5}$.

\begin{figure}
\setlength{\abovecaptionskip}{0.cm}
\setlength{\belowcaptionskip}{-0.5cm}
    \centering
    \includegraphics[width=3.20in]{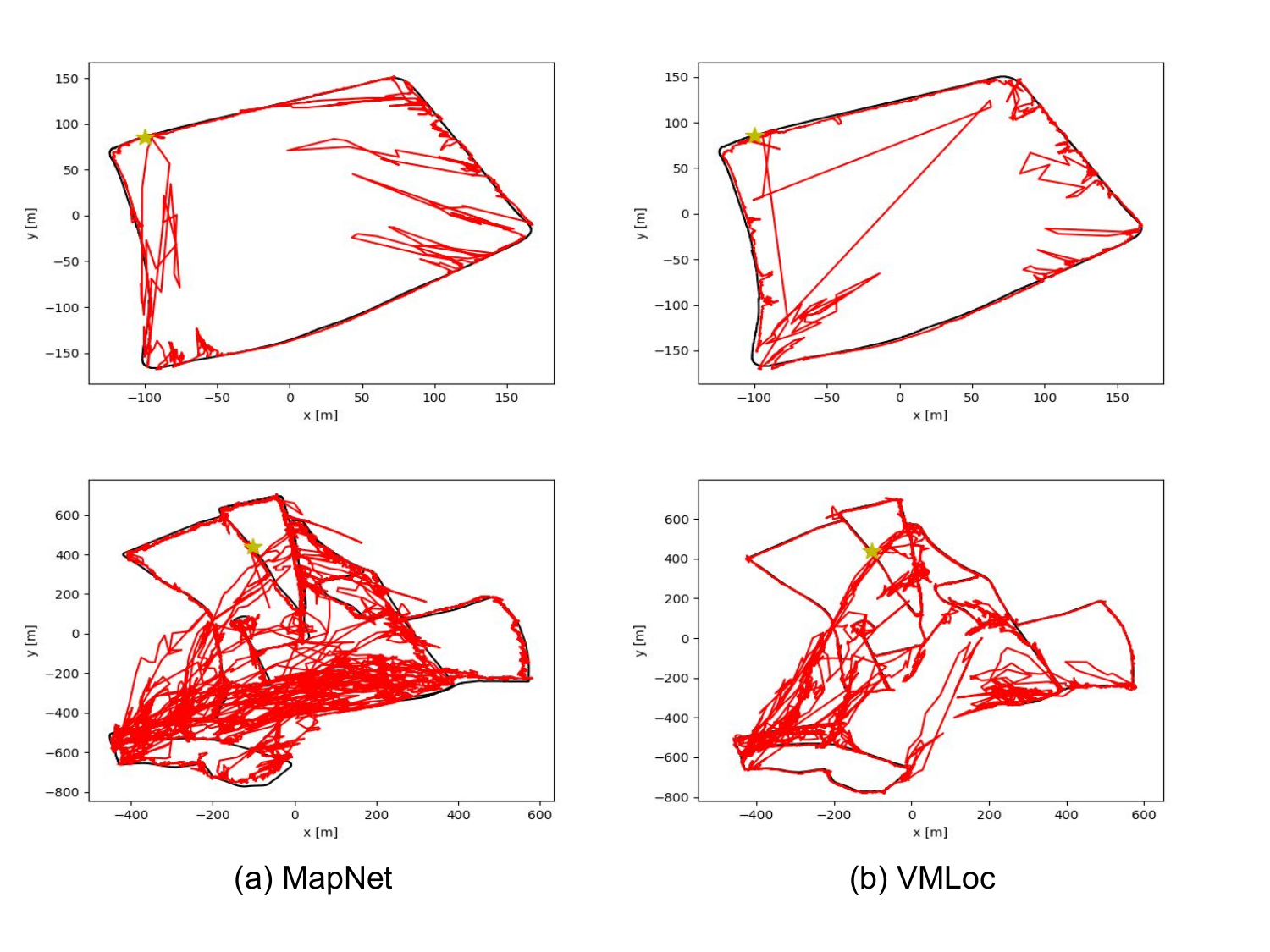}
    \caption{The generated trajectories of LOOP1 (Top) and FULL1 (Bottom) with proposed VMLoc (b) and the baselines MapNet (a). The yellow star denotes the starting point. The ground truth trajectories are shown in black lines, while the red lines are the predicted trajectories.}
    \label{fig:fig3}
\end{figure}

\begin{table*}[htbp]
\setlength{\tabcolsep}{10pt} 
\setlength{\abovecaptionskip}{0.cm}
\setlength{\belowcaptionskip}{-0.3cm}
\caption{The camera localization results from 7-Scenes Dataset. We report the median error of the position and orientation. Hourglass, PoseNet+, MapNet, and Atloc are only based on RGB image (\textbf{V}), while VMLoc and VidLoc use RGB and depth images (\textbf{V,D}).} 
% or only using the RGB image Without Temporal Constraints
\begin{center}
\scalebox{0.85}{\begin{tabular}{cccccccc}
\hline
\hline
\small
\textbf{Scene} & \textbf{Hourglass(V)} & \textbf{PoseNet+(V)} &\textbf{VidLoc(V,D)} & \textbf{MapNet(V)} &  \textbf{AtLoc(V)} & \textbf{VMLoc(V,D)}\\ 
\hline
Chess & $0.15m, 6.17^{\circ}$ & $0.13m, 4.48^{\circ}$ & $0.16 m, \text{NA}$ & $0.08 m, 3.25^{\circ}$  &  $0.10 m, 4.07^{\circ}$ & $\textbf{0.10m},\textbf{3.70}^{\circ}$ \\
Fire & $0.27m, 10.8^{\circ}$ & $0.27m, 11.3^{\circ}$ & $0.19 m, \text{NA}$ & $0.27 m, 11.7^{\circ}$ &  $0.25 m, 11.4^{\circ}$ & $\textbf{0.25m},\textbf{10.5}^{\circ}$ \\
Heads & $0.19m, 11.6^{\circ}$ & $0.17m, 13.0^{\circ}$ & $0.13 m, \text{NA}$ & $0.18 m, 13.2^{\circ}$  &  $0.16 m, 11.8^{\circ}$ & $\textbf{0.15m},\textbf{10.8}^{\circ}$ \\
Office & $0.21m, 8.48^{\circ}$ & $0.19m, 5.55^{\circ}$ & $0.24 m, \text{NA}$ & $0.17 m, 5.15^{\circ}$ &  $0.17 m, 5.34^{\circ}$ & $\textbf{0.16m}, \textbf{5.08}^{\circ}$ \\
Pumpkin & $0.25m, 7.01^{\circ}$ & $0.26m, 4.75^{\circ}$ &$0.33 m, \text{NA}$& $0.22 m, 4.02^{\circ}$  &  $0.21 m, 4.37^{\circ}$ & $\textbf{0.20m}, \textbf{4.01}^{\circ}$ \\
Red Kitchen & $0.27m, 10.2^{\circ}$ & $0.23m, 5.35^{\circ}$ &$0.28 m, \text{NA}$  & $0.23 m, 4.93^{\circ}$  &  $0.23 m , 5.42^{\circ}$ & $\textbf{0.21m},\textbf{5.01}^{\circ}$ \\
Stairs & $0.29m, 12.5^{\circ}$ & $0.35m, 12.4^{\circ}$ &$0.24 m, \text{NA}$ & $0.30 m, 12.1^{\circ}$  &  $0.26 m, 10.5^{\circ}$ & $\textbf{0.24m},\textbf{10.0}^{\circ}$ \\
\hline
Average & $0.23m, 9.53^{\circ}$ & $0.23m, 8.12^{\circ}$ &$0.23 m, \text{NA}$& $0.21 m, 7.77^{\circ}$  &  $0.20 m, 7.56^{\circ}$ & $\textbf{0.19m},\textbf{7.01}^{\circ}$ \\
\hline
\end{tabular}}
\label{table:Both_7SCENE}
\end{center}
\end{table*}

\begin{table}[htbp]
\setlength{\tabcolsep}{10pt} 
\setlength{\abovecaptionskip}{0.cm}
\setlength{\belowcaptionskip}{-0.3cm}
\caption{The camera localization results for Oxford RobotCar Dataset. We report the mean error of the position and orientation for MapNet, AtLoc, and VMLoc.} 
\begin{center}
\scalebox{0.85}{\begin{tabular}{ccccccc}
\hline
\hline
\small
\textbf{Scene} & \textbf{MapNet(V)} &  \textbf{AtLoc(V)} & \textbf{VMLoc(V,D)}\\ 
\hline
LOOP1 & $8.76 m, 3.46^{\circ}$ & $8.61 m, 4.58^{\circ}$ & $\textbf{7.70m}, \textbf{3.23}^{\circ}$  \\
LOOP2 & $9.84 m, 3.96^{\circ}$ & $8.86 m, 4.67^{\circ}$ & $\textbf{7.76m}, \textbf{3.16}^{\circ}$\\

FULL1 & $41.4 m, 12.5^{\circ}$ & $29.6 m, 12.4^{\circ}$  & $\textbf{19.5m}, \textbf{4.32}^{\circ}$\\
FULL2 & $59.3 m, 14.8^{\circ}$ & $48.2 m, 11.1^{\circ}$ & $\textbf{35.2m}, \textbf{8.99}^{\circ}$ \\
\hline
Average & $29.8 m, 8.68^{\circ}$ & $23.8 m, 8.19^{\circ}$ & $\textbf{17.5m}, \textbf{4.92}^{\circ}$ \\
\hline
\end{tabular}}
\label{table:Both_OX}
\end{center}
\end{table}

\section{Experiments}
\subsection{Datasets} 
Our proposed VMLoc framework is evaluated on two common public datasets: 7-Scenes \cite{shotton2013scene} and Oxford RobotCar \cite{maddern20171}. 
\textbf{7-Scenes Dataset} was collected by a Kinect device, consisting of RGB-D image sequences from seven indoor scenarios. The ground truth was calculated by KinectFusion algorithm. We split the data as training and testing set according to the official instruction.
\textbf{Oxford RobotCar Dataset} contains multimodal data from car-mounted sensors, e.g., cameras, lidar, and GPS/IMU. We use the same data split of this dataset named LOOP and FULL as in \cite{brahmbhatt2018geometry} and \cite{wang2019atloc}. As there is no depth map in the original Oxford Robot Car dataset, the depth map is obtained by projecting the sparse lidar to the RGB image as shown in Figure \ref{fig:fig1}.

\subsection{Training Details}
Our approach is implemented by using PyTorch. The model is trained and tested with an NVIDIA Titan V GPU. During the training process, both RGB images and depth maps are taken as the input, which are rescaled with the shortest side in the length of 256 pixels and normalized into the range of $[-1,1]$. In the case of VMLoc, the sampling number $k$ is set to be 10. The batch size is set to be 64 and the Adam optimizer is used in the optimization process with the learning rate $5\times 10^{-5}$ and the weight decay rate $5\times 10^{-5}$. The training dropout rate is set to be $0.5$ and the initialization balance weights are $\beta_0 = -3.0$ and $\gamma_0 =0.0$. All experiments have been conducted for 5 times to guarantee the reproducibility.

\subsection{Baseline and Ablation Study}
We use the recent state-of-the-art pose regression models as our baseline. Five representative learning-based models, i.e. Hourglass \cite{melekhov2017image}, PoseNet+ \cite{kendall2017geometric}, VidLoc \cite{clark2017vidloc}, MapNet \cite{brahmbhatt2018geometry}, and AtLoc \cite{wang2019atloc} are compared with our proposed method. %AtLoc reported the state-of-the-art performance in the 7-scenes and Oxford RobotCar datasets. 
AtLoc depends on single-image to realize accurate pose estimation. MapNet uses a sequence of images for localization which generally performs better than single image localization. We compared with MapNet to show the high accuracy achieved by our single image localization algorithm.

To prove the effectiveness of each module in VMLoc, we compare our model with the image-VMLoc which takes a single RGB image as its input while keeping the importance weighting and the unbiased objective function. To verify the performance of our fusion mechanism, we compare VMLoc with attention-VMLoc which uses attention without PoE and importance weighting, %and PoE-VMLoc which discards the attention mechanism while keeping the PoE distribution fusion mechanism. 
and PoE-VMLoc which uses the MVAE to fuse different distribution without using importance weighting and proposed unbiased objective function. Finally, to prove the robustness of our model, we test the performance of our model under different data degradation conditions.

\begin{table}[htbp]
\setlength{\tabcolsep}{4pt} 
\setlength{\abovecaptionskip}{0.cm}
\setlength{\belowcaptionskip}{-0.3cm}
\caption{The ablation study of camera localization results with Oxford RobotCar Dataset. We report the mean error of the position and orientation for image-VMLoc, attention-VMLoc, PoE-VMLoc and VMLoc.}
\begin{center}
\scalebox{0.85}{\begin{tabular}{ccccc}
\hline
\hline
\textbf{Scene} & \tabincell{c}{\textbf{image}\\ \textbf{VMLoc}} & \tabincell{c}{\textbf{attention}\\ \textbf{VMLoc}} & \tabincell{c}{\textbf{PoE}\\ \textbf{VMLoc}} & \textbf{VMLoc}\\ 
\hline
LOOP1 & $8.60 m, 4.57^{\circ}$ & $9.16 m, 4.96^{\circ}$ & $8.57m, 3.98^{\circ}$ & $\textbf{7.70m}, \textbf{3.23}^{\circ}$  \\
LOOP2 & $8.50 m, 3.90^{\circ}$ & $9.78 m, 5.66^{\circ}$ & $ 8.99m, 3.79^{\circ}$ & $\textbf{7.76m}, \textbf{3.16}^{\circ}$\\
FULL1 & $30.1m, 10.8^{\circ}$ & $31.2 m, 6.04^{\circ}$ & $30.0 m, 7.54^{\circ}$ & $\textbf{19.5m}, \textbf{4.32}^{\circ}$\\
FULL2 & $48.1 m, 9.61^{\circ}$ & $46.5 m, 10.1^{\circ}$ & $45.9 m, 10.5^{\circ}$ & $\textbf{35.2m}, \textbf{8.99}^{\circ}$ \\
\hline
Average & $23.9 m, 7.22^{\circ}$ & $19.3 m, 5.35^{\circ}$ & $18.7 m, 5.16^{\circ}$ & $\textbf{17.5m}, \textbf{4.92}^{\circ}$ \\
\hline
\end{tabular}}
\label{table:Ablation}
\end{center}
\end{table} 
%\vspace{-1cm}

\begin{table}[htbp]
\setlength{\tabcolsep}{1.5pt} 
\setlength{\abovecaptionskip}{0.cm}
\setlength{\belowcaptionskip}{-0.3cm}
\caption{The robustness study against corrupted input in Oxford RobotCar Dataset.}
\begin{center}
\setlength{\abovecaptionskip}{0.cm}
\setlength{\belowcaptionskip}{-0.3cm}
%\scalebox{0.9}{
\scalebox{0.85}{\begin{tabular}{ccccc}
\hline
\hline
\textbf{Corrupt.} &
\tabincell{c}{\textbf{AtLoc(V)}} & \tabincell{c}{\textbf{attention}\\ \textbf{VMLoc}} & \tabincell{c}{\textbf{PoE}\\ \textbf{VMLoc}} & \textbf{VMLoc}\\ 
\hline
\tabincell{c}{No} & $48.2 m, 11.1^{\circ}$ & $46.1 m, 9.50^{\circ}$ & $45.8 m, 11.50^{\circ}$ & $\textbf{35.2m}, \textbf{8.99}^{\circ}$  \\
\hline
\noalign{\smallskip}
\multicolumn{5}{c}{\textit{RGB}} \\
\hline
\tabincell{c}{$lvl=1$} & $314.3 m, 53.1^{\circ}$ & $\textbf{229.1 m}, \textbf{46.7}^{\circ}$ & $ 235.5m, 41.5^{\circ}$ & $241.9m, 48.9^{\circ}$  \\
\hline
\tabincell{c}{$lvl=2$} & $-$ & $484.7 m, 85.8^{\circ}$ & $478.0 m, 85.2^{\circ}$ & $\textbf{464.7m}, \textbf{85.5}^{\circ}$\\
\hline
\noalign{\smallskip}
\multicolumn{5}{c}{\textit{LIDAR}} \\
\hline
\tabincell{c}{$lvl=1$} & $-$ & $47.9 m, 9.6^{\circ}$ & $49.7 m, 12.1^{\circ}$ & $\textbf{36.5m}, \textbf{8.95}^{\circ}$\\
\hline
\tabincell{c}{$lvl=2$} & $-$ & $53.9 m, 9.29^{\circ}$ & $87.1 m, 36.3^{\circ}$ & $\textbf{38.7m}, \textbf{9.70}^{\circ}$ \\
\hline
\end{tabular}} 
\label{table:Robust}
\end{center}
\end{table}

\subsection{The Performance of VMLoc in Indoor Scenarios}
We first evaluate our model on the 7-Scenes dataset to demonstrate its effectiveness in fusing RGB and depth data in indoor scenarios. Table \ref{table:Both_7SCENE} shows the comparison between VMLoc and the competing approaches. The results are reported in the median error (m). VMLoc and VidLoc use both RGB images and depth data, while others are based on RGB images only. Our proposed VMLoc outperforms the other five baseline algorithms in terms of both the position and orientation error. Compared to AtLoc, %, the previous state-of-the-art method based on a single image, 
 VMLoc shows a $5.0\%$ improvement upon the position accuracy and a $7.3\%$ improvement in the rotation. In particular, in the Stairs scenario, VMLoc can reduces AtLoc position error from $0.26m$ to $0.24m$ and the rotation error from $10.5^{\circ}$ to $10.0^{\circ}$. This because Stairs is a highly texture-repetitive scenario and the structure information captured by the depth map can help to improve the network performance. The improvements made in this case match our expectations and are mainly because of the introduction of depth maps which provide external information for the pose estimation.

\subsection{Performance of VMLoc in the Driving Scenarios}
We further evaluate our models on the Oxford RobotCar outdoor dataset. Table \ref{table:Both_OX} summarizes the results of this experiments. Compared to AtLoc, VMLoc generates $26.5 \%$ and $40.0\%$ improvement in position and rotation accuracy respectively. We notice that the overall improvement is due to a large increase in rotation estimation as the projection of point cloud can better capture the structural information of the scene which can largely help estimating more accurate orientation. While the improvements in the position accuracy are less evident than that in rotation accuracy, the main reason may be that the point cloud in the Oxford RobotCar dataset is relatively sparse which can only provide limited complementary geometric information for position estimation. The visualization of MapNet and VMLoc trajectories for LOOP1 and FULL1 are shown in Figure \ref{fig:fig3}. VMLoc yields a closer prediction w.r.t. the ground truth and a smoother trajectory than MapNet.

\begin{figure}
\setlength{\abovecaptionskip}{0.cm}
\setlength{\belowcaptionskip}{-0.5 cm}
    \centering
    \includegraphics[width=3.20in]{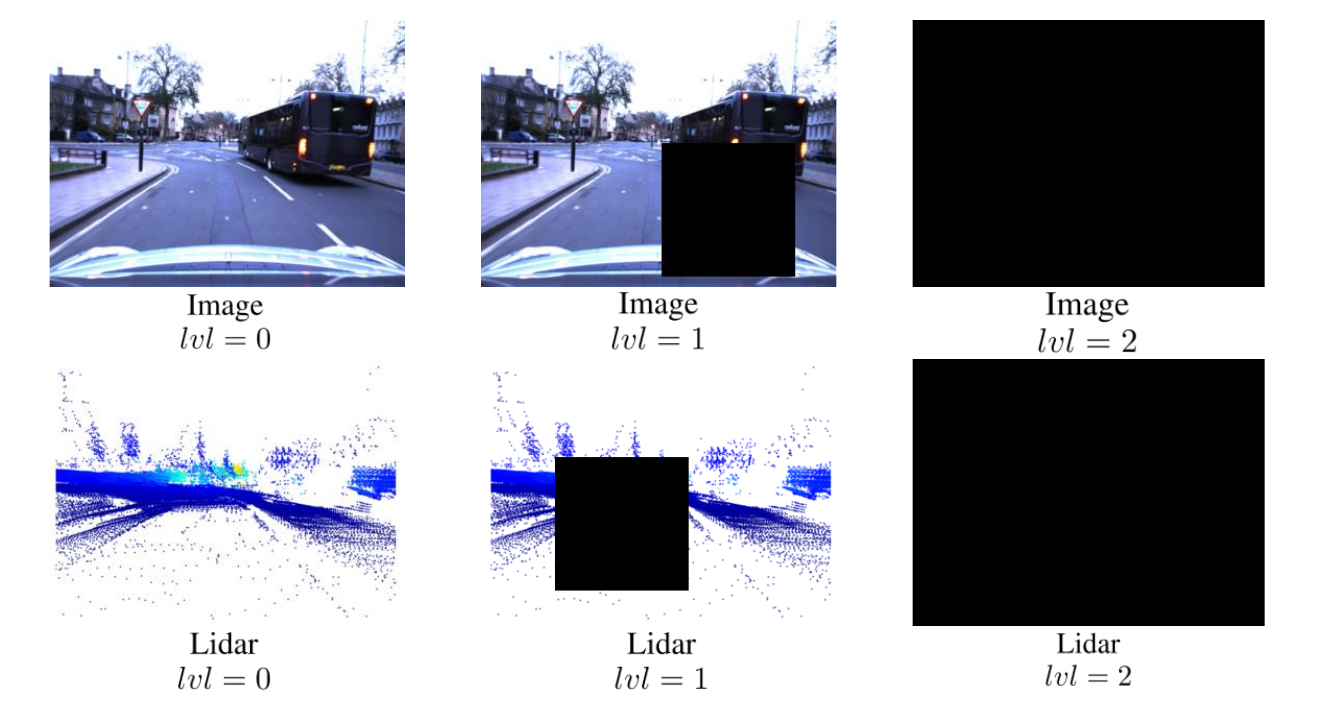}
    \caption{Input images and input projections of lidar with different levels of corruption.}
    \label{fig:fig4}
\end{figure}

\subsection{Ablation Studies of VMLoc}
To further verify the performance of the fusion mechanism, the ablation studies are conducted on Oxford Robot Car Dataset. The result is shown in Table \ref{table:Ablation}. With the help of variational learning and importance weighting, image-VMLoc performs better than MapNet and AtLoc as the variational learning learns the latent distribution of input module, which can better deal with the variation of different scenes. By directly concatenating the features of different modalities, attention-VMLoc does not provide a better performance which indicates that the concatenation is not an effective fusion module. By fusing RGB and depth features through PoE, PoE-VMLoc performs better than image-VMLoc, while the slight increase indicates that PoE-VMLoc also does not make full use of the lidar inputs. The best performances in all 4 scenes are achieved by VMLoc which indicates that it is a more effective fusion mechanism than attention-VMLoc and PoE-VMLoc. Moreover, compared with the performance in LOOP1 and LOOP2, a larger increase is yielded in FULL1 and FULL2. Since FULL1 and FULL2 are more complicated than LOOP1 and LOOP2 \cite{maddern20171}, the model cannot regress precise position only with the help of lidar.

\subsection{Computational Complexity}
The usage of importance weighting may raise concerns about the computational complexity of our algorithm. Nevertheless, during inference, VMLoc requires almost the same GPU time and the same FLOPS as attention-VMLoc and PoE-VMLoc. As the importance weighting is only used in the objective function in the training process, VMLoc has no appreciable difference compared with PoE-VMLoc during inference. With regards to the training process, when the importance weighting is set to 10 and VMLoc, attention-VMLoc and PoE-VMLoc share the same latent feature dimension of each module, $T_{VMLoc} \approx 1.23 T_{PoE} \approx 1.02 T_{attention}$  where $T$ represents the running time. We can see that the importance weighting does not greatly increase the computational complexity, as the importance sampling mainly influences the pose regressor part of VMLoc which only accounts for less than $5\%$ of the number of parameters.    

\subsection{Robustness Evaluation}
To demonstrate the robustness of VMLoc in the case of missing input, we test our models in the condition where input modules are degraded or missing. We adapt the data corruption method from \cite{chen2019selective}. In order to describe different data corruption conditions, we define the data corruption level from $lvl=0,1,2$. $lvl=0$ means that there is no corruption; $lvl=1$ means that there is mask of dimensions $128\times128$ pixels overlaying the input; $lvl=2$ means that the input is totally missing (see Figure \ref{fig:fig4}).

Table \ref{table:Robust} shows the performance of different fusion mechanisms under these corruption conditions. It is clear that when either lidar or image input is corrupted, the performances of all fusion mechanisms deteriorate. This indicates that all three learning-based fusion mechanisms make use of all inputs to regress the position and the deterioration of any input would affect their performance. However, in most cases, VMLoc performs better than attention-VMLoc, PoE-VMLoc, and AtLoc, which verifies the effectiveness of our fusion mechanism. 
Meanwhile, we also notice that the corruption of the image has a larger influence than the corruption of the lidar signal. This denotes that in this case, these fusion mechanisms still mainly relies on RGB image to make the decision which can be due to two reasons. On the one hand, the lidar signal in the Oxford Robot Car dataset is relatively sparse. On the other hand, even though the lidar signal has a wide viewpoint and is invariant to illumination, it is less informative \cite{tinchev2018seeing,tinchev2019learning}.

\section{Conclusion}
Effectively exploiting multimodal data for localization is a challenging problem due to the different characteristics among various sensor modalities. In this paper, we have proposed a novel multimodal localization framework (VMLoc) based on multimodal variational learning. In particular, we designed a new PoE fusion module by employing unbiased objective function based on importance weighting, which is aimed to learn the common latent space from different modalities. Our experiments have shown that this approach produces more accurate localization compared to existing single image or multimodal learning algorithms, either on benign conditions or when the input data are corrupted.

\bibliography{aaai}

\end{document}